# Generate Logical Equivalence Questions

Xinyu Wang, Haoming Yu, Yicheng Yang, Zhiyuan Li[*]

*Department of Computer Science*
*Beijing Normal-Hong Kong Baptist University*
Email: *goliathli@uic.edu.cn*

## ABSTRACT

Academic dishonesty is met with zero tolerance in higher education, yet plagiarism has become increasingly prevalent in the era of online teaching and learning. Automatic Question Generation (AQG) presents a potential solution to mitigate copying by creating unique questions for each student. Additionally, AQG can provide a vast array of practice questions.

Our AQG focuses on generating logical equivalence questions for Discrete Mathematics, a foundational course for first-year computer science students. A literature review reveals that existing AQGs for this type of question generate all propositions that meet user-defined constraints, resulting in inefficiencies and a lack of uniform question difficulty. To address this, we propose a new approach that defines logical equivalence questions using a formal language, translates this language into two sets of generation rules, and develops a linear-time algorithm for question generation.

We evaluated our AQG through two experiments. The first involved a group of students completing questions generated by our system. Statistical analysis shows that the accuracy of these questions is comparable to that of textbook questions. The second experiment assessed the number of steps required to solve our generated questions, textbook questions, and those generated by multiple large language models. The results indicated that the difficulty of our questions was similar to that of textbook questions, confirming the quality of our AQG.

[*]Corresponding Author

## 1 Introduction

In recent years, the growing use of online resources in academia has raised concerns about academic dishonesty. According to a report from China Education Daily, the number of plagiarism cases among college students has significantly increased since the outbreak of COVID-19 [1]. The report has indicated that the shift to online learning, combined with study pressure, has made cheating easier. In another survey conducted by China Youth Daily, 71.5% of college students admitted to witnessing or participating in academic misconduct, including plagiarism, during the pandemic [2]. This is a significant increase from the 52.3% reported by the same survey conducted before the pandemic. The survey has also found that 38.1% of students believe that online learning has increased the ease of academic misconduct. Another study [3] has reported an increase in plagiarism risk in Taiwan with the growth of online learning resources. One reason for the rise of plagiarism cases is the lack of invigilation during assessments. According to Xinhua News Agency, "the absence of a physical learning environment and the inability of teachers to monitor students' behavior in real time create a favorable environment for academic misconduct" [4]. The data show that online classes and the epidemic have significant impacted college students' tendencies toward plagiarism. As educators in Computer Science, we have zero tolerance for cheating.

The most effective way to minimize plagiarism is to give different questions to different students. If each student receives entirely unique questions, they cannot copy from each other. Thus, randomized question generation is considered.

There are usually two ways to randomize questions. One is to create a large manually designed question bank. When assignments or exam papers are created, different students are assigned to different sets of questions. However, this method becomes less effective as the student population is large. The question bank has to be large enough to minimize hash collisions, which creates a heavy workload for question designers. Additionally, questions need to be regularly updated to prevent leaks.

Another common approach is parameterizing questions using a piece of students' personal information, like student ID. For example, the original question is "Find the root of the quadratic function $f(x) = x^2 - 1$" which can be parameterized as "Find the root of the quadratic function $f(x) = x^2 - a$, where $a$ is the last digit of your student ID". Nevertheless, this method is only applicable to questions with numerical constants in courses like Calculus or Linear Algebra. But for proof problems in Discrete Mathematics, this method is less effective. Proofs, different from calculations, do not have any constant to be parameterized. For example, "Prove propositional equivalence of $p \equiv p \vee (p \wedge q)$".

Thus, this research aims to design a question generator for Discrete Math with the following requirements:

R1. different students receive different questions;
R2. students cannot obtain new questions through regeneration;
R3. generated questions have similar difficulty levels; and
R4. the difficulty level can be controlled.

Currently, we focus on questions regarding logical equivalences. Students are given two expressions in propositional logic and are required to show their equivalence using equivalence laws. We choose this question type because it is covered in the first chapter of Discrete Mathematics. Our long-term goal is to generalize the generation method to various question types across multiple fields of mathematics and computer science. Thus, an additional requirement is

R5. our generator has the potential to be generalized to all question types.

Furthermore, from an application perspective,

R6. our generator runs in linear time with respect to the question length.

Automatic Question Generation is an existing class of algorithms to generate questions, which has been studied by many scholars. Ahmed et al. [5] present an algorithm that generates questions and solutions for natural deductions using a data structure called Universal Proof Graph, which encodes all possible inference rules over all small propositions. Other approaches leverage existing proof assistants. Mostafavi et al. [6] design a system based on Deep Thought [7], a logic proof tutor, while Oliveira et al. [8] use Alloy [9], a modeling language for program synthesis, to generate propositional calculus questions.

However, these existing approaches are not tractive for our scenario. These existing frameworks generate questions by enumerating all valid propositions, which grow exponentially many with respect to the proposition length. Then, to ensure different students receive different questions of similar difficulty, an exhaustive hardness test on all generated questions is required, making the process very inefficient. Therefore, our AQG must use random seeds, which purely depends on the students' information. In addition, our approach must incorporate a user-defined difficulty classification at each generation step. Difficulty control is infeasible with existing frameworks because the implementation of generation is sealed within them. Also, for generalization purposes, logical equivalence questions must be regarded as a formal language, which is precisely specified by lexicon, syntax, and semantics. Consequently, each generated question must be a spelling correct, grammatically correct, and meaningful sentence within the language.

Large Language Models (LLMs) can also be candidates for question generation. We have tried GPT3.5, GPT4, and GPT4.o. The experiment shows that these LLMs are not robust. The generated questions are sometimes too easy, sometimes too difficult, and occasionally wrong. A detailed discussion is given in Section 4.

In the general domain of AQG for Mathematics or Computer Science, Kurdi et al. [10] present a systematic review of the results of research on the problem of automated generation for educational purposes. Singhal et al. [11] describe a framework that helps teachers quickly generate large numbers of questions on a geometry topic for high school mathematics. This method ensures the validity of the generated questions. Singhal and Henz [12] also introduce a framework to limit the scope of knowledge the question examines. The proposed method generates a region graph to represent regions formed by input geometry objects. Singhal et al. proposed a framework [13], which considers user-defined difficulty for AQG. Several other papers present a range of relevant frameworks as well. Polozov et al. [14] introduced how to make personalized problems based on individual students' characteristics and preferences. The system utilizes machine learning techniques and natural language processing to tailor the difficulty level, context, and topics to suit each student's learning needs. Wang [15] and Gupta [16] proposed some techniques to generate mathematical problems. Both of them utilize natural language processing. Zavala [17] and Thomas et al. [18] provided a good tool to generate programming exercises. Zavala focuses on using Semantic-Based Abstract Interpretation Guidance (AIG) to automatically generate programming exercises with varying levels of difficulty while Thomas presents a stochastic tree-based method for generating program-tracing practice questions that require learners to understand code execution flow. For more about AQG for computer science or mathematics questions, readers are referred to [19] [20] [21].

This paper is organized as follows. Section 2 provides preliminary information that is crucial for comprehending the entirety of our research. Section 3 proceeds to elaborate on the implementation process and the specific methodologies employed in this study. Section 4 presents experiment results to compare the performance of our generator with human-designed questions. Section 5 concludes this paper.

## 2 Preliminaries

A logical equivalence question gives two *propositions*. Students need to show that the two propositions are *logically equivalent* using *equivalence laws*. A proposition can be *atomic* or *compound*. An atomic proposition is either a constant ($True$ or $False$) or a propositional variable (represented by English letters in lower case). A compound proposition can be a *unary operator* followed by a proposition, or two propositions connected by a *binary operator*. In this paper, we only discuss the unary operator *negation* ¬, and the binary operators *conjunction* ∧, *disjunction* ∨, *implication* →, and *biconditional* ↔. The *precedence* of these operators is shown in *Table 1*, where a smaller value represents a higher precedence.

| Operator | Precedence |
|---|---|
| ¬ | 1 |
| ∧ | 2 |
| ∨ | 3 |
| → | 4 |
| ↔ | 5 |

*Table 1. Precedence of logical operators*

Two propositions $(P, Q)$ are *equivalent*, denoted as $P \equiv Q$, if $P$ and $Q$ have the same truth value for any possible assignment of propositional variables. Other than drawing a truth table, logical equivalences can also be proven using the following 21 laws.

| | | |
|---|---|---|
| **Identity** | $p \wedge T \equiv p$ | $p \vee F \equiv p$ |
| **Domination** | $p \wedge F \equiv F$ | $p \vee T \equiv T$ |
| **Commutative** | $p \wedge q \equiv q \wedge p$ | $p \vee q \equiv q \vee p$ |
| **Idempotent** | $p \vee p \equiv p$ | $p \wedge p \equiv p$ |
| **Negation** | $p \wedge \neg p \equiv F$ | $p \vee \neg p \equiv T$ |
| **Absorption** | $p \vee (p \wedge q) \equiv p$ | $p \wedge (p \vee q) \equiv p$ |
| **Associative** | $(p \wedge q) \wedge r \equiv p \wedge (q \wedge r)$ | |
| | $(p \vee q) \vee r \equiv p \vee (q \vee r)$ | |
| **De morgan** | $\neg (p \wedge q) \equiv \neg p \vee \neg q$ | |
| | $\neg (p \vee q) \equiv \neg p \wedge \neg q$ | |
| **Double negation** | $\neg \neg p \equiv p$ | |
| **Implication** | $p \to q \equiv \neg p \vee q$ | |
| **Bi-Implication** | $p \leftrightarrow q \equiv (p \to q) \wedge (q \to p)$ | |
| **Distributive** | $p \vee (q \wedge r) \equiv (p \vee q) \wedge (p \vee r)$ | |
| | $p \wedge (q \vee r) \equiv (p \wedge q) \vee (p \wedge r)$ | |

To specify a *formal language*, we need to define
- the *alphabet* $\Sigma$, a set of symbols;
- the set of *tokens* $\mathcal{T}$, which are *regular expressions* with names;
- the *syntax* $\Gamma$, which is in Backus-Naur form; and
- the *semantic* $\mathcal{C}$, which is a set of propositions with names.

$\Gamma$ is a set of production rules in the form $V ::= a_1 a_2 \cdots a_n$. $V$ and $a_1 \cdots a_n$ are symbols of $\Gamma$. Symbols appear on the left-hand side of "::=" are non-terminals. Symbols only appear on the right-hand side are terminals. Terminals are tokens in $\mathcal{T}$. A *substitution* involves replacing the occurrence of $V$ in a string by "$a_1 a_2 \cdots a_n$", denoted as $\ldots V \ldots \Rightarrow \ldots a_1 a_2 \cdots a_n \ldots$. A *production* is a sequence of substitutions, which starts from the start non-terminal $S$ and ends with a string of terminals. Moreover, the entire process of the generation can be represented by a *syntax tree*, where non-terminals are internal vertices, terminals are leaves, and vertex $V$ has children $a_1, a_2, \cdots, a_n$ if the rule $V ::= a_1 a_2 \cdots a_n$ is used in one step of a production.

An *attribute system* of a formal language consists of a set of *attribute definitions* for all non-terminals and terminals. The attribute of a terminal depends on the terminal itself, which call *intrinsic*. The attribute definition of a non-terminal $V$ is associated with the production rules in which $V$ is involved. Assume $V ::= a_1 a_2 \cdots a_n$ is a production rule, the attribute definition $V.at \coloneqq$ $f(a_1.at, a_2.at, \cdots a_n.at)$ means that the attribute of $V$ is defined by the attributes of $a_1 \cdots a_n$, which is called *synthesized*. If the attribute definition is $a_1.at \coloneqq f(V.at, a_2.at, \cdots, a_n.at)$, the attribute is *inherited*. If all attributes of terminals and non-terminals can be computed by a single Depth-First search on the syntax tree, the language is *L-attributed*. In this project, attribute definitions are used to translate syntax trees into expressions.

Readers are referred to [22] for more details of propositional logic and logical equivalences, and [23] for more about the syntax and semantics of a formal language.

## 3 Method

Our AQG takes a piece of students' information and a set of parameters, which control the difficulty of generated questions, from instructors as input and produces two equivalent propositions. The generation consists of 3 phases, random seed generation, syntax tree generation, and propositional expression construction.

### 3.1 Randomness

The first phase converts students' information into MD5 codes as random seeds, which guarantees a low probability of hash collision [24]. In consequence, if the randomness of question generation purely depends on the random seed, different students can hardly get the same question. Furthermore, using random seeds allows our algorithm to generate the same set of problems for a student multiple times. Then, students cannot cherry-pick questions by smashing the "generate" button.

In detail, an MD5 code contains 16 hexadecimal digits. Each digit will be used to make a pseudo-random decision one by one. Once all 16 digits are used once, the circulation on the MD5 code starts again from the first digit but with an offset. If the offset is a prime number, the sequence of digits can hardly repeat.

| Rounds | Hex code | Offset |
|---|---|---|
| 1 | $39cf0c951da2210198e0db94f91a4b3a$ | 0 |
| 2 | $3c091a209ed9f1439fc5d21180b49aba$ | 1 |
| 3 | $3f9d21ebfa3905a190994acc1208d41b$ | 2 |
| ⋮ | ⋮ | ⋮ |

*Table 2 An example of the circulation on the MD5 digits*

### 3.2 Syntax tree generation

The second phase is the main phase of our AQG. It takes the random seed as input and generates two correlated syntax trees. The propositions presented by the syntax trees are equivalent. Before introducing the detailed generation scheme, the "language" of propositional equivalence is formally defined as below.

- $\Sigma = \{\vee, \wedge, \to, \neg, T, F, (,)\} \cup \{A, \cdots, Z\}$;
- $\mathcal{T} = \{\ bop: \vee\ |\ \wedge\ |\ \to, uop: \neg, con: T\ |\ F, par: (\ |\ ),$
  $var: [A - Z]^+$ except $T, F\}$;
- $\Gamma = \{\ I ::= uop\ E\ |\ E\ bop\ E,$
  $E ::= con\ |\ var\ \}$;
- $\mathcal{C} = \{\ Ab \vee$: if $p, q$ are propositions, then $p \vee (p \wedge q) \equiv p$,
  $Ab \wedge$: if $p, q$ are propositions, then $p \wedge (p \vee q) \equiv q$,
  $\cdots\ \}$.

In the set of tokens $\mathcal{T}$, $bop, uop, con,$ and $var$ are the names of tokens, meaning binary operators, unary operators, Boolean constants, and Boolean variables respectively. And similarly, in $\mathcal{C}$, $Ab \vee$ and $Ab \wedge$ are the names of semantic propositions. In this language, we only define the semantic of proposition equivalences. This semantic definition is called algebraic semantic, which is a set of algebraic properties specifying the meaning of the equivalence symbol " $\equiv$ ". Each specification is in First-Order language, meaning that it may contain some propositional variables. For example, $p$ is a variable in $p \vee (p \wedge q) \equiv p$. In this language definition, the behaviors of operators are not defined in computational semantic or denotational semantic because the behaviors are not concerned by our AQG. In addition, we only list two semantic rules in this definition because of the page limit. Readers should understand that there 21 semantic rules in total, one for each propositional equivalence law.

Furthermore, parentheses are omitted and the grammar is ambiguous. For example, $T \vee F \wedge T$ can be understood in two different ways: $(T \vee F) \wedge T$ or $T \vee (F \wedge T)$. However, this is not an issue for our project. Our goal is to generate some grammatically correct propositions, but not analyze the structure of given expressions. Precedence is already presented by the hierarchy of syntax trees. And parentheses will be added when syntax trees are converted to expressions.

Next, generation rules are defined in the shape of Backus-Naur form. Note that two propositions are equivalent in two ways: structural equivalent - in the same syntactic structures; and semantic equivalent - in the same meaning but with different structures. Therefore, generation rules are of two kinds: $\mathcal{R}_1$ consists of syntactic generations, while $\mathcal{R}_2$ consists of semantic generations. To maintain the equivalence, two propositions are generated simultaneously. Thus, generation rules are paired in groups of two. The first rule in each group defines the generation on the first proposition, while the second rule is for the second proposition. Furthermore, some steps of generation intend to build multiple substructures. So, a generation rule consists of a sequence of multiple production rules. In detail,

|  | Name | First proposition | Second proposition |
|---|---|---|---|
| $\mathcal{R}_1$ | $r_{lit}$ | $E_i ::= lit_j$ | $E_i' ::= lit_t'$ |
|  | $r_\wedge$ | $E_i ::= E_j \wedge E_k$ | $E_i' ::= E_j' \wedge E_k'$ |
|  | $r_\vee$ | $E_i ::= E_j \vee E_k$ | $E_i' ::= E_j' \vee E_k'$ |
|  | $r_\rightarrow$ | $E_i ::= E_j \rightarrow E_k$ | $E_i' ::= E_j' \rightarrow E_k'$ |
|  | $r_\neg$ | $E_i ::= \neg E_j$ | $E_i' ::= \neg E_j'$ |
| $\mathcal{R}_2$ | $r_{Ab\vee}$ | $E_i ::= E_j \vee E_h$ $E_h ::= E_j \wedge E_k$ | $E_i' ::= E_j'$ |
|  | $r_{Ab\wedge}$ | $E_i ::= E_j \wedge E_h$ $E_h ::= E_j \wedge E_k$ | $E_i' ::= E_j'$ |

The prime symbol ' intends to show that $E_i'$ is the copy of $E_i$, but $E_i$ is in the first syntax tree, while $E_i'$ is in the second. They produce the same substructures in syntax trees. The index $i$ of $E_i$ is to distinguish different $E$ 's during the generation. Similar to the semantic $\mathcal{C}$, $\mathcal{R}_2$ has 21 semantic generation rules. But only 2 absorption rules are listed to save space. Also, each rule is given a name for further reference. For example,

| Step | Proposition 1 | Proposition 2 | Rule |
|---|---|---|---|
| 1 | $E_0 \Rightarrow E_1 \vee E_2$ $\Rightarrow E_1 \vee E_1 \wedge E_3$ | $E_0' \Rightarrow E_1'$ | $r_{Ab\vee}$ |
| 2 | $\Rightarrow lit_1 \vee lit_1 \wedge E_2$ | $\Rightarrow lit_1'$ | $r_{lit}$ |
| 3 | $\Rightarrow lit_1 \vee lit_1 \wedge lit_2$ | $\Rightarrow lit_1'$ | $r_{lit}$ |

The corresponding syntax tree is given in Figure 1.

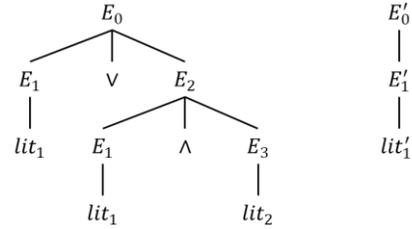

*Figure 1. The two syntax trees for a generation*

With the above generation rules, the syntax tree generation is presented in Algorithm 1.

**Algorithm 1.** SyntaxTreeGeneration $(C, \mathcal{R}_1, \mathcal{R}_2, V, p_r, l, m)$
**Input:** $C = c_1 c_2 \cdots c_{16}$ is an MD5 code;
$\mathcal{R}_1$ and $\mathcal{R}_2$ are the two sets of rules;
$V$ is a non-terminal in $\mathcal{R}_1 \cup \mathcal{R}_2$;
$p_r$ is the current probability;
$l$ is the desired number of iterations;
$m$ is the maximum number of iterations.
**Output:** two syntax trees $T_1$, $T_2$
1. $c_1 = getHex(C)$
2. $c_2 = getHex(C)$
3. **if** $l \geq m$ **then**
4.    Use the rule $E ::= lit$ on every non-terminal.
5.    And terminal the generation.
6. **else if** $\frac{c_1}{16} < p_r$ **then**
7.    Roulette select a rule $r$ in $\mathcal{R}_1$ using $\frac{c_2}{16}$.
8.    Assume $r$ is $(V ::= a_1 \cdots a_n,\ V' ::= a_1' \cdots a_n')$ and $V'$ is the copy of $V$ in $T_2$.
9.    Construct a subtree rooted at $V$ with children $a_1, \cdots, a_n$ in $T_1$.
10.    Do the similar thing to $V'$ in $T_2$.
11.    If $a_i$ is a non-terminal $E_k$, the index $k$ is a new index which has not been used by this generation.
12.    $p_r = p_r + p_c$
13.    **for each** non-terminal $E_k$
14.       SyntaxTreeGeneration$(C, \mathcal{R}_1, \mathcal{R}_2, V, p_r, l + 1, m)$
15. **else**
16.    Reset $p_r$.
17.    Does the similar thing but using the rules from $\mathcal{R}_2$.
18. **endif**

This is a recursive algorithm, where $p_r$, $p_c$, and $m$ are user-defined parameters that control the question difficulty. The argument $V$ is initialized as the initial variable $E_0$ in the first function call. The $getHex$ function (defined in Section 3.1) is called twice to cyclically pick two hexadecimal digits from the MD5 code $C$. The first hex $c_1$ decides whether the algorithm selects a rule from $\mathcal{R}_1$ to create a syntactic structure or a rule from $\mathcal{R}_2$ to apply an equivalence law. Line 7 uses a roulette selection with a probability $\frac{c_2}{16}$ to decide which rule is selected in $\mathcal{R}_1$. The algorithm uses a guaranteed probability to ensure that at least one rule in $\mathcal{R}_2$ is applied every few iterations of the recursion. If the algorithm applies a rule in $\mathcal{R}_1$, $p_r$ is increased by $p_c$ on line 12, which is a user-defined probability increment; otherwise, $p_r$ is reset to its initial value on line 16. The value $l$ is initialized as 0 and tracks the number of iterations. If $l$ exceeds $m$, the maximum desired number of iterations, the generation directly terminates by applying $E_i ::= lit_j$ on every non-terminal $E_i$.

## 3.3 Logical expression construction

The last phase constructs the two logical expressions from the syntax trees. This phase involves the establishment of an attribute system to ascertain the interpretation of each node in the syntax tree. The attribute system defines two attributes for each symbol, "$exp$" - the expression produced by the grammar symbol and "$pre$" the precedence of the last performed operator in the expression, which is for proper parenthesizing of the logical expression. We only want to add the necessary parentheses. For example, the brackets in $(p \lor q) \land s$ are not avoidable. The attribute "$pre$" is defined for this purpose.

First, intrinsic attributes are defined for terminals. For each operator, the expression is the token itself and its precedence is defined in *Table 1*. Taking ∨ as an example,
$$\lor.exp ::= \lor \text{ and } \lor.pre ::= 3$$
The expression of a literal is a variable name that follows the regular expression $var$ (in $\mathcal{T}$). Our generator simply constructs a valid variable name randomly. The precedence of a literal is undefined and assigned to be ∞ for computation convenience. Thus,
$$lit_i.exp ::= Random() \text{ and } lit.pre ::= \infty$$
Remember that $lit'_i$ is the duplication of $lit_i$. Thus, we also need
$$lit'_i.exp ::= lit_i.exp \text{ and } lit'_i.pre ::= \infty$$

---
if $(E_k.pre < \lor.pre)$ then
   if $(E_j.pre < \lor.pre)$ then
     $E_i.exp ::= (E_j.exp) \land ((E_j.exp) \lor (E_k.exp))$
   else if $(E_j.pre < \land.pre \text{ and } E_j.pre > \lor.pre)$ then
     $E_i.exp ::= (E_j.exp) \land (E_j.exp \lor (E_k.exp))$
   else
     $E_i.exp ::= E_j.exp \land (E_j.exp \lor (E_k.exp))$
else
   if $(E_j.pre < \lor.pre)$ then
     $E_i.exp ::= (E_j.exp) \land ((E_j.exp) \lor E_k.exp)$
   else if $(E_j.pre < \land.pre \text{ and } E_j.pre > \lor.pre)$ then
     $E_i.exp ::= (E_j.exp) \land (E_j.exp \lor E_k.exp)$
   else $E_i.exp ::= E_j.exp \land (E_j.exp \lor E_k.exp)$
$E'_i.exp ::= E_j.exp$
$E_i.pre ::= \land.pre$
$E'_i.pre ::= E_j.pre$

---

*Figure 2. Attribute definition for the generation rule $r_{Ab\lor}$*

For non-terminals, let $r: (E_i ::= a_1 \cdots a_n, E'_i ::= a'_1 \cdots a'_m)$ be a generation rule. In general, the logical expression of $E_i$ is the concatenation of $a_1.exp, \cdots, a_n.exp$ (and similar for $E'_i$). Exceptions happen when operators are involved. If a binary operator is contained, the generation rule must be in the shape $E_i ::= E_j \text{ op } E_k$. Then, if the precedence of $E_j$ (or $E_k$) lower than the precedence of $op$, then the expression of $E_j$ (or $E_k$) must be included by a pair of parentheses. The unary operator "¬" is treated in the same way. And the precedence $E_i.pre$ tracks the precedence of the last performed operator in $E_i$. Figure 2 shows the attribute definition for $r_{Ab\lor}$ as an example.

With the attribute definitions, the logical expression construction algorithm is straightforward. It is simply a DFS traversal implemented as a recursion. See Algorithm 2 for details.

**Algorithm 2.** Expression ($V$)
```
Input: A syntax tree T rooted at vertex V
Output: V.exp, the logical expression of T
1.  if V is a leaf and a literal then
2.      V.exp = Random()
3.      V.pre = ∞
4.  else if V is a leaf and an operator then
5.      V.exp = V
6.      V.pre = the precedence of operator V
7.  else
8.      for each child U_i from U_1 to U_n of V
9.          Expression(U_i)
10.     endfor
11.     Compute V.exp, V.pre by V's attribute
        definition
12. endif
```

Combining Algorithm 1 and Algorithm 2, our generator clearly runs in linear time regarding the proposition length.

## 3.4 Difficulty control

The syntax tree generated by phase 2 purely depends on the MD5 encoding of students' information. In consequence, some unlucky students may receive harder questions than others. To unify the difficulties, the syntax tree generation works with a difficulty control, which concerns three following issues
- the minimum length $m$ of propositions,
- the number of equivalence laws $e$ used during the production, and
- the reasoning difficulty of applying the equivalence laws.

The first two issues can easily be measured and controlled. From Algorithm 1, the generation simply halts once the length of propositions reaches $m$. And the number of equivalence laws $e$ is in fact controlled by the guarantee probability increment $p_c$. In our experiment, we let $m = 5$ and $p_c = \frac{1}{8}$. Results show that these values can generate solid questions.

The last issue is relatively subjective and cannot be easily measured. From the historical feedback of assignments in our institution, we observe that applying some equivalence laws is harder to students than other equivalence laws. For example, students can usually easily apply the Identity law $p \lor F \equiv p$, but difficultly apply the Absorption law $p \lor (p \land q) \equiv p$. Following this intuition, the equivalence laws are manually classified into three categories:

- "Easy" - Identity, Double Negation, Domination;
- "Median" - De Morgan, Distributive, Idempotent, Negation;
- "Hard" - Absorption, Commutative, Associative, Implication, Bi-Implication.

Commutative and Associative laws are usually considered trivial. But in our design, another equivalence law, which can be Absorption, Idempotent, Double Negation, or Identity, is always cooccurred with Commutative or Associative. This technique makes Commutative and Associative hard. In Algorithm 1, line 17 circulants on "Median", "Hard", and "Easy". Every time when the algorithm selects a rule from $\mathcal{R}_2$, it will first pick a median rule, then a hard rule, then an easy rule, and go back to a median rule again. This design intends to select median rules as much as possible, and also take more hard rules than easy ones.

## 4 Experiment

In this project, we accomplish two rounds of experiments. The first experiment tests the quality and difficulty of problems generated by our algorithm and validates whether they are comparable to human-designed problems. 52 year-one students from our college have participated in the experiment. Most of them were assigned 3 generated questions and few were given 4 questions. Students were required to finish the questions in 30 minutes. The test is open book. Students have access to the equivalence laws, which simulates exactly the same environment when students are doing homework assignments. Students' answers are marked by instructors and classified into 3 categories: "correct", "half-correct", and "incorrect", which receive 100%, 50%, and 0% respectively. Then, the correct rate is measured by the average marks of all answers. At the same time, we collect the historical students' performance on logical equivalence questions from homework assignments over 4 academic years, 2021, 2022, 2023, and 2024. Data from quizzes in 2021 is also collected. The questions given to students of those semesters are textbook questions from [22] [25]. Simple statistics and comparisons show that the average grade of generated questions lies among the historical average. (See Table 3. for details.) Thus, we are confident that the generated questions can serve as good practices for students without being too frustrating.

| Student Batch | Number of Participants | Average grade |
| --- | --- | --- |
| Our Experiment | 52 (164 questions) | 65.24% |
| 2021 Spring | 179 | 68.33% (Assignment) 66.1% (Quiz) |
| 2022 Spring | 236 | 51.78% (Assignment) |
| 2023 Spring | 194 | 60.98% (Assignment) |
| 2024 Spring | 280 | 66.45% (Assignment) |

*Table 3. Experiment Result*

Another experiment is a "white box" testing. It selects 30 questions from textbooks and generates 60 questions using our AQG, ChatGPT3.5, and ChatGPT4.o. We analyze the question difficulty by determining the minimum number of steps required to solve them, with one step representing the application of one equivalence law. For GPT generation, we use the prompt "Give me 60 extremely hard logical equivalence questions". The analysis shows that 10 (out of 30) of the textbook questions need 4 steps to solve, while the remaining textbook questions are ranged from 1 to 9 steps. Our AQG generated questions are primarily centered around 3 or 4 steps. In contrast, the GPT generated questions were more varied. 52 (out of 60) ChatGPT3.5 generated questions are unique, with 27 of them being correct (the two propositions are indeed equivalent). As for ChatGPT4.o, it generates 49 distinct questions, and only 20 are correct. This comparison demonstrates that our AQG can effectively control the question difficulty. Detailed experiment data is given in Table 4.

| Steps | 1 | 2 | 3 | 4 | 5 | 6 | 7 | 8 | 9 | 10 | >10 | NEQ | Total |
| --- | --- | --- | --- | --- | --- | --- | --- | --- | --- | --- | --- | --- | --- |
| AQG | 1 | 6 | 31 | 21 | 1 | 0 | 0 | 0 | 0 | 0 | 0 | 0 | 60 |
| Textbook | 2 | 1 | 4 | 10 | 6 | 2 | 1 | 2 | 2 | 0 | 0 | 0 | 30 |
| ChatGPT3.5 | 10 | 4 | 3 | 5 | 1 | 1 | 0 | 1 | 1 | 0 | 1 | 25 | 52 |
| ChatGPT4.o | 0 | 4 | 2 | 4 | 3 | 0 | 1 | 0 | 1 | 2 | 3 | 29 | 49 |

*Table 4. The number of steps required to solve the questions. The column ">10" means that the number of steps is greater than 10. And "NEQ" means that the question is wrong. The two propositions are not equivalent.*

## 5 Conclusion and Future Works

In this paper, we provide a framework for a new automatic generation of propositional logical questions with difficulty control based on the user input. Our system is able to generate distinct questions for different students efficiently. Experiment results provide solid evidence for the feasibility of our generator, affirming that our approach has the potential to produce homework assignments with appropriate difficulty and quality.

One major future work is to generalize the generator from logical equivalence to all question types. From Section 3.2, logical equivalence questions are described as a formal language, and this formal language is converted to a generator. Thus, a generalization can be done on this procedure. A universal generator can take a question type definition (language definition), then construct a question generator (like what this paper has done) for the specific question type.

Another future work can be the design of an auto-grader. Once our generator is applied in real teaching and learning evaluation, tones of questions will be released to students. Human markers become infeasible to grade all these questions. To make grading efficient, an auto-grader must be designed.